%% file: main.tex
\pdfoutput=1

\documentclass[11pt]{article}
\input{sections/00_decl}

\title{\method: Generative Marmoset Spoken Language Modeling}

\author{Talia Sternberg\textsuperscript{1}, Mickey London\textsuperscript{2}, David Omer\textsuperscript{2}$^{*}$, Yossi Adi\textsuperscript{1}\thanks{Equal-contribution last authors}\\ 
 \textsuperscript{1}The School of Computer Science and Engineering\\
 \textsuperscript{2}The Edmond and Lily Safra center for Brain Sciences (ELSC)\\
 Hebrew University of Jerusalem, Israel\\ \texttt{talia.sternberg@mail.huji.ac.il}}

\begin{document}
\maketitle
\begin{abstract}

Marmoset monkeys exhibit complex vocal communication, challenging the view that nonhuman primates' vocal communication is entirely innate, and show similar features of human speech, such as vocal labeling of others and turn-taking. Studying their vocal communication offers a unique opportunity to link it with brain activity—especially given the difficulty of accessing the human brain in speech and language research. Since Marmosets communicate primarily through vocalizations, applying standard LLM approaches is not straightforward. We introduce \emph{Generative Marmoset Spoken Language Modeling} (\method), an optimized spoken language model pipeline for Marmoset vocal communication. We designed a novel zero-shot evaluation metrics using unsupervised in-the-wild data, alongside weakly labeled conversational data, to assess \method and  demonstrate its advantage over a basic human-speech-based baseline. \method generated vocalizations closely matched real resynthesized samples acoustically and performed well on downstream tasks. Despite being fully unsupervised, \method effectively distinguish real from artificial conversations and  may support further investigations of the neural basis of vocal communication and provides a practical framework linking vocalization and brain activity. We believe \method stands to benefit future work in neuroscience, bioacoustics, and evolutionary biology. Samples are provided under: \href{https://pages.cs.huji.ac.il/adiyoss-lab/GmSLM/}{this link}. 

\end{abstract}

\input{sections/01_intro}
\input{sections/03_background}
\input{sections/04_approach}

\input{sections/05_setup}
\input{sections/06_results}
\input{sections/02_related}

\input{sections/07_conclusion}

\section*{Acknowledgments}
This research was funded by the Center for Interdisciplinary Data Science Research (CIDR) grant number $3035000503$, Israeli Science Foundation (ISF) grant numbers $2049/22$ and $1331/23$, and The Gatsby Charitable Foundation.

\section*{Ethical Statement}
Since the proposed approach involves encoding, modeling, and decoding \monks vocalizations, careful attention must be given to data collection. All data collection procedures were conducted with the necessary ethical committee approvals, ensuring that the well-being of the \monks was maintained to the highest standards.

\bibliography{refs}

\clearpage
\appendix
\section{Appendix}
\label{sec:appendix}
\subsection{Pre-processing Details} \label{appendix:preprocessing}

We develop a multi-stage pipeline aimed at precisely segmenting \monk calls while effectively reducing environmental noise contamination. 

We applied a high-pass filter at 5 kHz, as \monk calls primarily occur above this frequency, while lower frequencies mostly consist of noise or human speech. A spectrogram was generated using a hop length of $512$ and a window size of $2048$, allowing for frame-wise filtering in the frequency domain. Individual frames were classified as noise based on the following algorithm. We begin by filtering time-frequency bins with energy below a predefined threshold. Next, individual frames are identified as noise candidates based on their filtered time-frequency variance and density. These noise candidates are then aggregated over time, and segments are classified as noise if their duration is either less than $0.5$ seconds or greater than $2$ seconds. A similar analysis is then performed on the remaining audio, keeping only segments whose durations fall within the typical marmoset call range of $[0.25, 4]$ seconds. Finally, call boundaries are determined by the first and last frames of each detected segment. This method results in a robust call segmentation while reducing false detections, achieving a precision of $0.975$ and a recall of $0.78$ (computed on a held-out manually labeled set).

\subsection{Dataset}
\label{sec:data_setup}
Audio segments from the preprocessing stage were downsampled to $16$ kHz, ensuring the necessary frequency range for detecting \monk vocalizations while preserving call integrity. The dataset was then divided into training ($80\%$), validation ($10\%$), and test ($10\%$) sets. Each segment was $10$ seconds long and contained as many \monk calls as possible. The average call length was $0.8$ seconds, with approximately $3.3$ calls per segment. In total, the dataset comprises $\sim360$ hours of audio, with approximately $40\%$ consisting of detected \monk calls and the remaining $60\%$ comprising inter-call gaps, which may still include background noise. Overall, the dataset includes $216K$ samples. Dataset statistics are detailed in Table \ref{tab:general data}.

\begin{table}[t!]
\centering
\small
\setlength{\tabcolsep}{5pt} 
\renewcommand{\arraystretch}{1.2} 
\begin{tabular}{lcccc}
\toprule
 & Train & Test & Validation & Total \\
\midrule
N.~Samples & 173,328 & 21,666 & 21,666 & 216,660 \\
Total Hours & 292 & 36 & 37 & 365 \\
Call Hours & 125 & 15 & 15 & 155 \\
\bottomrule
\end{tabular}
\caption{Dataset statistics after preprocessing and filtering. The dataset includes detected marmoset calls and inter-call gaps.\label{tab:general data}}
\end{table}

For \dataft, the dataset consists of three one-hour-long audio recordings at $16$kHz, capturing interactions among four different \monk. A total of $56$ call-response pairs were identified, in which one \monk calls another, and the recipient responds, with gaps of up to $10$ seconds between calls. Using these pairs, we generated approximately $600$ augmented sequences to create the evaluation task, as outlined in Section~\ref{sec:eval_method}.

\subsection{Training Configuration}
\label{sec:train_conf}
For all HuBERT configurations, we used the \texttt{hubert-base} model with $12$ Transformer layers. This model encodes raw audio into $768$-dimensional frame representations, generating one frame every $20$ms. The masking span is set to  $l = 10$, with  $p = 8\%$  of encoder output frames masked. Optimization is performed using Adam \cite{kingma2014adam} with  $\beta = (0.9, 0.98)$, applying linear warm-up of the learning rate to $0.0035$ over the first $3\%$ of steps, followed by a linear decay to zero. Training is conducted for $150K$ steps on four $24GB$ GPUs, with a batch size of up to $32$ seconds of audio.

Clustering was conducted using the MiniBatchKMeans function from scikit-learn, utilizing $50\%$ of the training and validation data, with a mini-batch size of $10K$ frames and k-means++ initialization with $20$ random starts. All experiments are implemented using the \textit{fairseq} library \citep{ott2019fairseq}. Inference for the first stage, leveraging the pre-trained speech model, was performed using \textit{textlesslib}~\citep{kharitonov2022textless}.

For the uLM, we use a vanilla Transformer model as implemented in \textit{fairseq}~\cite{ott2019fairseq}. Specifically, we adopt the \texttt{transformer\_lm\_big} architecture, which consists of $12$ layers, $16$ attention heads, an embedding size of $1024$, a feed-forward network (FFN) size of $4096$, and a dropout probability of $0.1$. The model is trained as a causal language model (LM) on sequences of pseudotext units, with each sample containing up to $3,072$ units. The model is trained on $4$ GPUs for $50K$ steps with a batch size of up to $700$ samples. We use the \texttt{inverse\_sqrt} learning rate scheduler with $4k$ warm-up steps, starting with an initial learning rate of \(1\mathrm{e}{-7}\) and reaching a peak learning rate of $5e-45$. Optimization is performed using Adam  with \(\beta = (0.9, 0.98)\).
For generation, we apply a beam-search sampling, with beam-size of $5$ and a temperature of $1.5$.

For U2V, we use a modified version of the HiFi-GAN neural vocoder ~\cite{kong2020hifi}, following the adaptation by \citet{polyak21_interspeech} for unit-to-waveform conversion. The unit vocoder is trained on three $24GB$ GPUs for $400K$ steps.

\subsection{FAD groups}
\label{appendix:FAD}
The FAD scores across different audio comparisons (the lower the better). Different types of audio subsets, were compared to the same Original group (A) , together with modified versions of another subset of the original data (B), with no intersection between them.
Resynthesized refers to reconstructed versions of B, Generated includes vocalizations generated using first $3$ seconds prompts from B, Reversed contains temporally inverted sequences of B, and Random corresponds to Gaussian noise. All groups consist of $2$K examples, each lasting more than $5$ seconds.

\subsection{Classifier Training Configuration}
\label{appendix:classifier-details}
To evaluate the generalization capability of the model, we trained lightweight classifiers on the \datalab dataset for both vocalization type and caller identity prediction (see Section~\ref{sec:generalization}). Below, we provide implementation details and training setup for these classifiers.
We randomly split the labeled dataset into $90\%$ training and $10\%$ validation sets, ensuring the splits were balanced with respect to the class labels (either call type or caller identity). Each classifier was trained for $20$ epochs using a single GPU with $24GB$ memory. The models were optimized using the Adam optimizer combined with a polynomial learning rate scheduler.
The classifier takes as input pooled statistical features derived from the given representations outputs — specifically, the mean and variance computed over the temporal dimension of the representations, which are concatenated to form the input vector. This input is passed through three fully connected layers of decreasing size, each followed by layer normalization and ReLU activation functions. The final layer is a linear projection that outputs the logits for classification.

\begin{table}[t!]
\centering
\small
\setlength{\tabcolsep}{5pt} 
\renewcommand{\arraystretch}{1.2} 
\begin{tabular}{lccc}
\toprule
Model & Recall & Precision & F1 \\
\midrule
\textbf{GmSLM} & \textbf{91.96} & \textbf{90.14} & \textbf{90.72} \\
Marmoset HuBERT & 85.66 & 84.15 & 84.89 \\
Speech HuBERT & 88.88 & 83.60 & 86.15 \\

\bottomrule
\end{tabular}
\caption{Comparison of performance on the vocalization type classification task using different representations. GmSLM uses the final uLM layer, while HuBERT and Speech HuBERT use representations from the 9th layer.}
\label{tab:classifier_gmslsm_hubert}
\end{table}

\subsection{LSTM training configuration}
\label{appendix:LSTM-details}
For the lstm, we used a model as implemented in \textit{fairseq}~\cite{ott2019fairseq}. Specifically, we adopt the \texttt{lstm\_lm} architecture, consists of a single-layer unidirectional LSTM decoder with $512$-dimensional embeddings and hidden states, followed by a linear projection to the token vocabulary and a dropout probability of $0.2$
similar to the ulm transormer training, the model is trained as a causal language model (LM) on sequences of pseudotext units, with each sample containing up to $3,072$ units. The model is trained on $4$ GPUs for $50K$ steps with a batch size of up to $700$ samples. We use the \texttt{inverse\_sqrt} learning rate scheduler with $4k$ warm-up steps, starting with an initial learning rate of \(1\mathrm{e}{-7}\) and reaching a peak learning rate of $5e-45$. Optimization is performed using Adam  with \(\beta = (0.9, 0.98)\).
For generation, we apply a beam-search sampling, with beam-size of $5$ and a temperature of $1.5$.

\subsection{\monk Filterbank}
\label{appendix:monk_filterbank}
Marmoset vocalizations exhibit concentrated energy in the 5--8\,kHz range. Although MFCCs were originally designed for speech signals, they have become a standard representation for general audio tasks and are widely used across various fields. However, due to the acoustic mismatch between human speech and \monk vocalizations, we explored a feature representation better aligned with the \monk specific characteristics.
To this end, we constructed a custom filterbank focused on the 5--8,kHz band. This filterbank maintains the same dimensionality as the MFCCs, replaces the Mel scale with a linear frequency scale, and was used to train a uLM from scratch.

\begin{table}[t!]
\centering
\small
\setlength{\tabcolsep}{5pt} 
\renewcommand{\arraystretch}{1.2} 
\begin{tabular}{lccc}
\toprule
Model & Shuffle & Concat & Reversal \\
\midrule
5--8\,kHz Filterbank & 84.85 & 59.52 & 90.01 \\
MFCC + uLM & 82.30 & 60.70 & 89.90 \\
GmSLM (ours) & \textbf{84.84} & \textbf{79.94} & \textbf{90.45} \\
\bottomrule
\end{tabular}
\caption{Comparison of different input feature representations for uLM training. The 5--8\,kHz filterbank is designed to match the dominant frequency range of marmoset vocalizations.}
\label{tab:filterbank_comparison}
\end{table}

As shown in Table~\ref{tab:filterbank_comparison}, the 5--8\,kHz filterbank performs similarly to MFCCs, with a slight improvement on the Shuffle task. Nonetheless, GmSLM, which leverages HuBERT representations, achieves the best overall results.

\subsection{More Advanced Modeling Components}
Our pipeline currently relies on relatively dated components, including HuBERT and legacy vocoders. In recent years, the field has seen significant advancements in open-source speech and language models. For example, Whisper ~\cite{radford2022robustspeechrecognitionlargescale} and BEATs ~\citet{chen2022beatsaudiopretrainingacoustic} demonstrate strong performance in speech understanding, while modern pre-trained language models like LLaMA and Qwen offer improved capabilities over earlier LLMs ~\cite{qwen2025qwen25technicalreport,touvron2023llamaopenefficientfoundation}. 
However, applying these advanced tools in our setting poses specific challenges. Whisper is a fully supervised ASR model, which assumes the availability of written language—unsuitable in the case of Marmosets, where no such orthographic form exists. Similarly, BEATs is primarily optimized for audio classification, limiting its effectiveness in generative or sequence modeling tasks. By contrast, self-supervised learning (SSL) models, such as HuBERT, are better suited for modeling spoken language without written form. This approach aligns with standard practices in prior work on speech language models ~\cite{on_generative, twist,nguyen2024spiritlminterleavedspoken}.
We additionally explored the use of more recent language models by replacing the uLM with a pre-trained Qwen2.5-0.5B model, fine-tuned using the TWIST~\cite{twist} initialization strategy. For a controlled comparison, we trained two variants with the same architecture: (1) \textbf{GSLM}, where the Qwen model is trained from scratch, and (2) \textbf{TWIST}, where the Qwen model is initialized from pre-trained weights and fine-tuned on speech representations. Performance results are summarized in Table~\ref{tab:qwen-results}. Both approaches achieve comparable results across all evaluation tasks.
Finally, while it is true that a more advanced vocoder could improve the quality of audio output, our goal is not to maximize audio fidelity but to analyze the linguistic structure of uLM predictions. Thus, we leave the development of a more sophisticated vocoder to future work.

\begin{table}[t!]
\centering
\small
\setlength{\tabcolsep}{5pt} 
\renewcommand{\arraystretch}{1.2} 
\begin{tabular}{lccc}
\toprule
Model & Shuffle & Concat & Reversal \\
\midrule
TWIST & 84.88 & 68.21 & 88.78 \\
GSLM & 85.90 & 69.23 & 89.90 \\
\bottomrule
\end{tabular}
\caption{Evaluation results for Qwen-based uLMs trained from scratch (GSLM) and fine-tuned (TWIST).}
\label{tab:qwen-results}
\end{table}

\subsection{Limitations}
While \method introduces effective evaluation proxies, a key limitation remains the overall system evaluation. Assessing such a complex model is particularly challenging in a fully unsupervised setting. Future research should explore more fine-grained, task-specific evaluation methods. Additionally, since our approach was primarily tested under a single recording condition, evaluating its performance in out-of-domain scenarios remains an important direction for future work.

\subsection{AI Tools Usage}
AI tools have been used to assist in fixing grammar mistakes and sentence paraphrasing. Additionally, AI tools have been partially used to enhance code implementations. However, the authors carefully reviewed all content, ensuring these tools were only used as supportive aids and in responsible manner.

\subsection{Additional Results}
See units distribution for different vocalization types and vice versa in Figures \ref{fig:labels_distribution_of_units}-\ref{fig:wide_plot}. Attention maps visualization in Figure~\ref{fig:attention} and spectrogram visualizations in Figure~\ref{fig:spectrograms}. Audio samples can be found under \href{https://pages.cs.huji.ac.il/adiyoss-lab/GmSLM/}{this link}.

\input{Figures/spect_gen_rsynt_examples}

\end{document}

%% file: sections/00_decl.tex
\usepackage[final]{acl}

\usepackage{times}
\usepackage{latexsym}
\usepackage{acronym}
\usepackage{amsmath}
\usepackage{amssymb}
\usepackage{mathtools}
\usepackage{amsthm}

\usepackage[T1]{fontenc}

\usepackage[utf8]{inputenc}

\usepackage{microtype}

\usepackage{inconsolata}

\usepackage{graphicx}

\usepackage{booktabs}
\DeclareUnicodeCharacter{05BF}{}
\usepackage{xspace}

\newtoggle{release}
\togglefalse{release}
\toggletrue{release}

\iftoggle{release}{
\newcommand{\didi}[1]{}
\newcommand{\mickey}[1]{}
\newcommand{\talia}[1]{}
\newcommand{\adios}[1]{}
}
{
\newcommand{\didi}[1]{{\color{magenta}D: #1}}
\newcommand{\mickey}[1]{{\color{purple}M: #1}}
\newcommand{\talia}[1]{{\color{blue}T: #1}}
\newcommand{\adios}[1]{{\color{teal}Y: #1}}
}

\newcommand{\method}{\textsc{GmSLM}\xspace}
\newcommand{\datapre}{\textsc{ColonyDB}\xspace}
\newcommand{\dataft}{\textsc{PheeDB}\xspace}
\newcommand{\datalab}{\textsc{InfantMarmosetsVox
}\xspace}

\newcommand{\monk}{Marmoset\xspace}
\newcommand{\monks}{Marmosets\xspace}
\acrodef{GSLM}{Generative Spoken Language Modeling}

\newcommand{\newpara}[1]{\vspace{0.01cm}\noindent \textbf{#1}}

\usepackage{booktabs}  
\usepackage{multirow} 
\usepackage{makecell}  
\usepackage{amsmath}
\usepackage{graphicx}  
\usepackage{subcaption} 
\usepackage{lipsum}    
\usepackage{subcaption}  
\usepackage{xcolor}

%% file: sections/01_intro.tex
\section{Introduction}
\label{sec:intro}

\monk monkeys are small primates with surprisingly complex vocal communication. Although it was once assumed that nonhuman primates' vocal communication is entirely innate and inflexible, recent studies show that \monks can learn new vocalizations and even label their conspecifics~\citep{oren_vocal_2024}. These findings open the door for advanced computational analyses powered by large language models—to systematically investigate and decode \monk vocalization patterns, as a novel approach to uncover the cognitive and evolutionary mechanisms behind vocal communication. However, as \monks use vocal communication, with no well-defined orthography, it is unclear how recent advancements in LLMs, typically operate over discrete tokens, could be effectively utilized. 

\begin{figure}[t!]
    \centering
    \includegraphics[width=\columnwidth]{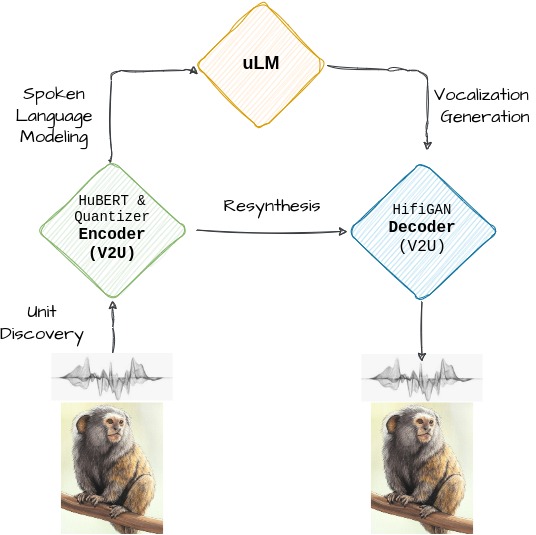}
    \caption{High level description of \method.}
    \label{fig:GMSLM}
\end{figure}

In recent years, the field of \emph{Generative Spoken Language Modeling} (GSLM), often known to as textless NLP, has gained traction~\cite{on_generative, twist}, demonstrating remarkable performance across various speech and audio tasks, encompassing both generative \citep{kharitonov2023speak, kreuk2022audiogen} and discriminative approaches \citep{chang2024speechprompt}. The objective of GSLM is to develop a spoken language model that operates without text. The typical GSLM framework begins with an unsupervised process for acoustic units discovery, producing a discrete representation of the audio signal. Then, a language model is applied to these units to estimate the likelihood of the audio signal. Finally, a generative network converts the units back into a time-domain audio signal. Since the GSLM pipeline enables efficient encoding, sequential modeling, and decoding of raw audio signals without relying on textual supervision, it could be ideally suited for modeling \monks vocalizations. 

In this work, we introduce \method (stands for Generative Marmoset Spoken Language Modeling), a tailored adaptation of the GSLM pipeline for \monks-vocalizations, where each modeling component is optimized specifically for \monk-vocalization. To evaluate \method, we developed a set of zero-shot evaluation metrics to measure the quality of its sequential modeling and conducted additional evaluations 
using weakly labeled conversational \monk data. A visual description can be seen on Figure~\ref{fig:GMSLM}.
We compare the proposed approach to the naive human-speech based baseline, and demonstrate \method significantly enhances system performance. Finally, we assess the quality of the sampled and generated \monks-vocalizations and empirically show these are comparable to their resynthesized counterparts. Leveraging manually annotated data (at the speaker level), we empirically demonstrate that, despite being entirely unsupervised, the proposed method assigns higher likelihood to authentic \monks conversations compared to unnatural ones. 
We evaluate \method on downstream supervised tasks, where it performs strongly, confirming its utility beyond the training objective. Ablation studies support the need for our model architecture given the data’s complexity. We also use \method to explore marmosets' vocal communication and interpret learned representations.
Overall, our work establishes a foundation for future research in animal language and related downstream applications. More broadly, it opens a new direction for unsupervised spoken language modeling in nonhuman species. We hope it inspires further research in this emerging field.

%% file: sections/03_background.tex
\section{Background}
\label{sec:back}

\subsection{Generative spoken language modeling}
The general \ac{GSLM} pipeline is comprised of three main modules: (i) Speech-to-unit, (ii) Unit language model, and (iii) Unit-to-speech, where each of these modules is trained separately. Speech resynthesis can be achieved while ignoring the language model and directly feeding the quantized units into the unit-to-speech module~\citep{polyak21_interspeech}. In the following paragraphs, we give detailed background for the three components mentioned above. 

\newpara{Speech-to-unit} module encodes the raw speech signal into a discrete representation. The common approach is first to encode the speech into a continuous representation and then quantize the representation to achieve a sequence of discrete units~\citep{tjandra2020transformer, polyak21_interspeech}.

Formally, denote the domain of audio samples by $\mathcal{X} \subset \mathbb{R}$. A raw signal is represented by a sequence of samples $x = (x_1,\ldots, x_T)$, where  $x_t\in\mathcal{X}$ for all $1\leq t \leq T$. Consider an encoder, $f$ that gets as input the speech utterance and produces low-frequency features $f(x) = (v_1, \ldots, v_{T'})$. \citet{on_generative}, evaluated several speech encoders, namely, Mel-spectrogram, wav2vec2~\citep{baevski2020wav2vec}, and HuBERT~\citep{hubert}. A k-means algorithm is applied over the models' outputs to generate discrete units, denoted as $z = (z_1,\ldots,z_{T'})$. Each element $z_i$ in $z$ is a positive integer, $z_i\in\{1,..,K\}$ for $1\le i \le T'$, $K$ is the number of discrete units.

\newpara{Unit language model} is trained on the unit sequences $z$, learning a distribution over them. This enables unconditional or conditional speech generation without text. Unlike text-based models, unit LMs can capture prosody~\citep{kharitonov2021text}, speaker identity~\citep{borsos2022audiolm}, or even natural dialogues~\citep{nguyen2022generative}.

\newpara{Unit-to-speech} module converts the speech discrete units to a raw waveform. \citet{on_generative} used a Tacotron2.0~\citep{tac} based model followed by WaveGlow~\citep{prenger2019waveglow} vocoder. Later, \citet{polyak21_interspeech} proposed a more efficient HiFi-GAN-based unit which convert units to speech directly. Further work~\citep{kreuk2021textless, lee2021direct} added emotional and duration modeling for tasks like emotion transfer and speech-to-speech translation

\subsection{Marmoset vocalization background}
 \monk monkey (Callithrix jacchus) are highly social primates that exhibit a wide repertoire of vocalizations~\citep{bezerra_structure_2008}. Their vocal production flexibility, including modulation of call features such as duration, intensity, complexity, and timing~\citep{brumm2004acoustic,eliades2012neural,roy2011vocal}, enable them to encode a wide range of social and emotional information~\citep{seyfarth2003signalers}, including callers' identity, sex and group affiliation,as well as group dialect, and receivers' identity \citep{norcross1993context, zurcher2017evidence,jones1993stability}.
 
 One particularly well-studied vocalization is the Phee call~\cite{chen_contact_2009}, a contact call ranging from $5.5$ to $10$kHz, which \monks is useed to engage in turn-taking dialogues with other conspecifics. Recent findings further suggest that \monks use the Phee call to encode the receiver’s identity, in a manner analogous to how humans use names to address each other~\cite{oren_vocal_2024}. These vocal characteristics  resemble human speech properties, including turn-taking in dialogue and vocal labeling of others ~\cite{oren_vocal_2024, osmanski2023perceptual}. Thus, \monks serve as a comparative model for vocal communication, offering insights for linguistics, cognitive science, and evolutionary biology.

\begin{figure}[t!]
    \centering
    \includegraphics[width=\columnwidth]{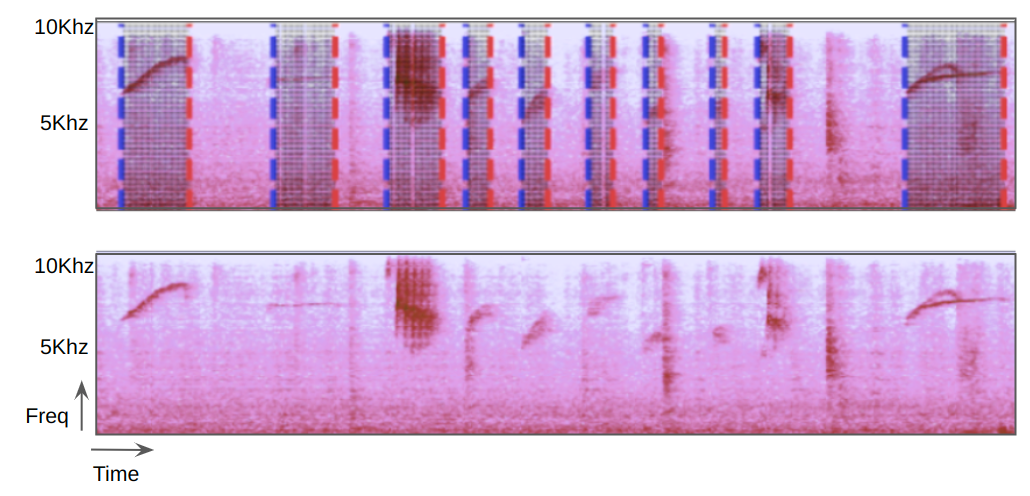}
    \caption{Segmentation example over an 18-second spectrogram from the original dataset. The x-axis represents time, while the y-axis represents frequency. As observed, the call bandwidth exceeds 5 kHz, with well-defined onset (blue lines) and offset (red lines) points computed in the preprocessing phase.\label{fig:spect}}
    \vspace{-0.3cm}
\end{figure}

%% file: sections/04_approach.tex
\section{Approach}
\label{sec:approach}
Recall that the proposed approach, (\method) builds on the \ac{GSLM} pipeline, which was originally designed for human speech. In this section, we outline the modifications and adaptations made to better suit Marmoset vocalizations. We begin by detailing the data collection and preprocessing phase, followed by an explanation of the adjusted system components. Finally, we describe the system evaluation process.

\subsection{Data}
\newpara{Dataset collection.} We utilize two distinct \monk-vocalization datasets. The first, referred to as \datapre, consists of unsupervised raw \monk vocalization recordings and is used for training \method. The second, called \dataft, is a smaller, weakly supervised dataset primarily used for model evaluation.

\datapre comprises continuous $24/7$ audio recordings of spontaneous, multi-speaker \monk vocalizations in a colony room. Monkeys, housed in family cages without visual contact but with vocal communication, produced diverse vocal exchanges. The dataset features a rich and diverse range of vocalizations exchanged naturally between different monkeys. Recordings were collected using an omnidirectional microphone positioned at the center of the colony room.

\dataft was collected in a controlled setup where monkey pairs, separated by a visual barrier, spontaneously engaged in \emph{Phee} call dialogues. Unlike colony recordings, these sessions captured structured turn-taking with a single vocalization type, and each caller's identity was fully labeled.

\newpara{Pre-processing.} We apply a multi-stage pipeline to segment \monk calls while minimizing environmental noise. A high-pass filter at $5$ kHz removes low-frequency noise, followed by spectrogram-based noise filtering and duration-based segmentation. This approach ensures robust call detection, achieving a precision of $0.975$ and recall of $0.78$. For a detailed explanation of the segmentation algorithm, see Appendix~\ref{appendix:preprocessing}. Fig~\ref{fig:data_stats} provides statistics on the pre-processed data, while a visual representation of the spectrogram is shown in Fig~\ref{fig:spect}.

\begin{figure}[t!]
    \centering
    \includegraphics[width=\linewidth]{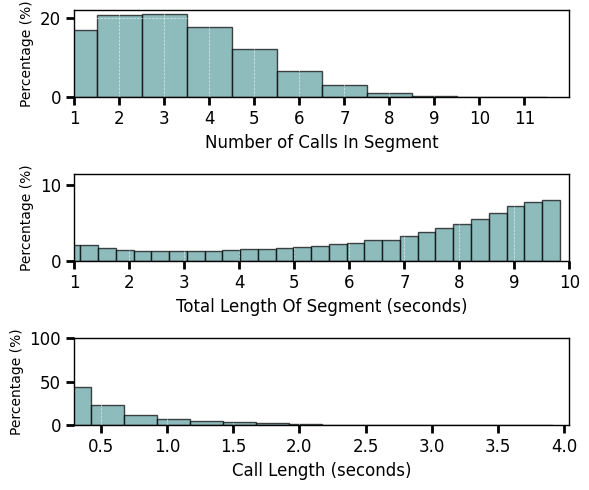}
    \caption{Statistics of the dataset after pre-processing. On average, each segment contains three calls, with an approximately equal distribution ranging from one to five calls. The total segment length is predominantly greater than $7$ seconds, while the mean call duration is approximately $0.8$ seconds.}
    \label{fig:data_stats}
\end{figure}

\begin{figure*}[t!]
    \centering
    \vspace{-1cm}
    \begin{subfigure}{0.24\textwidth}
        \centering
        \includegraphics[width=\textwidth]{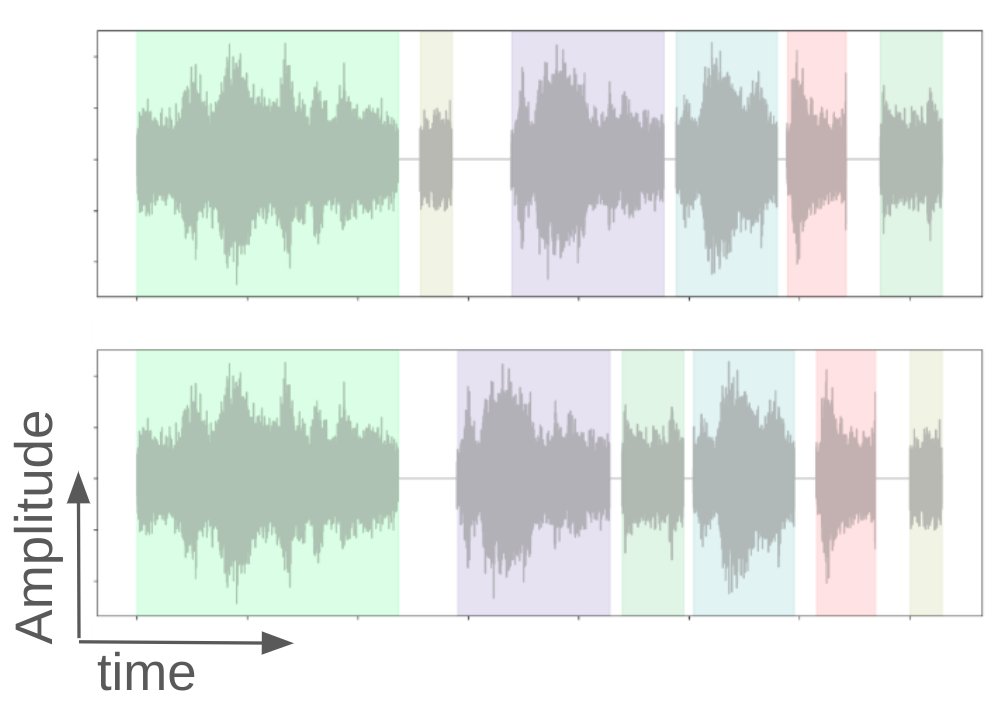}
        \caption{Shuffle}
        \label{fig:shuffle}
    \end{subfigure}
    \begin{subfigure}{0.24\textwidth}
        \centering
        \includegraphics[width=\textwidth]{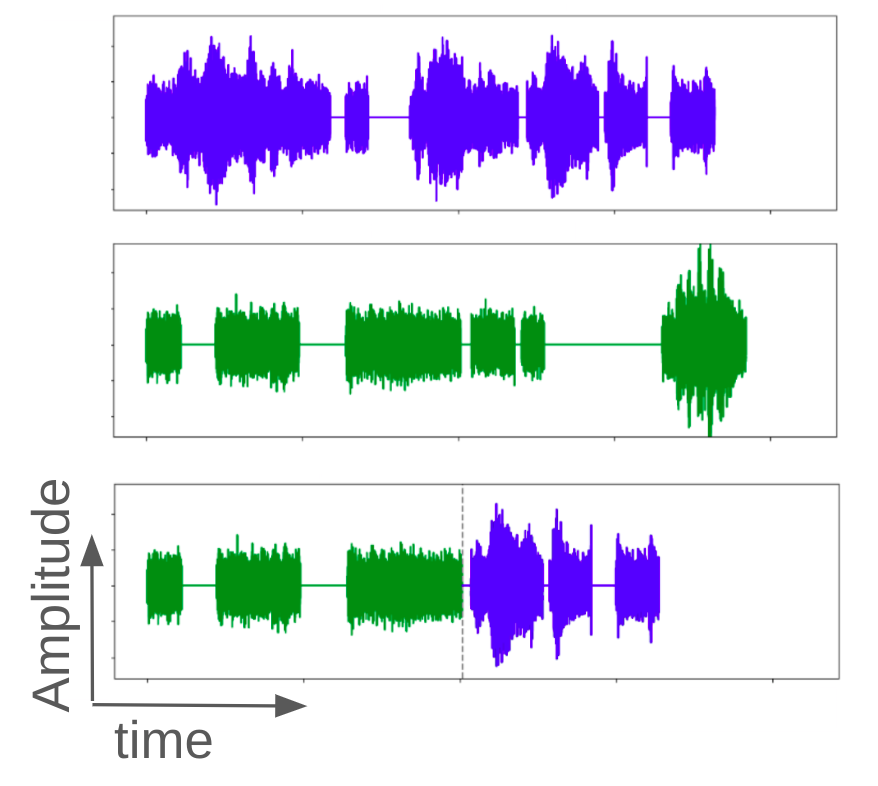}
        \caption{Concat}
        \label{fig:concat}
    \end{subfigure}
    \begin{subfigure}{0.24\textwidth}
        \centering
        \includegraphics[width=\textwidth]{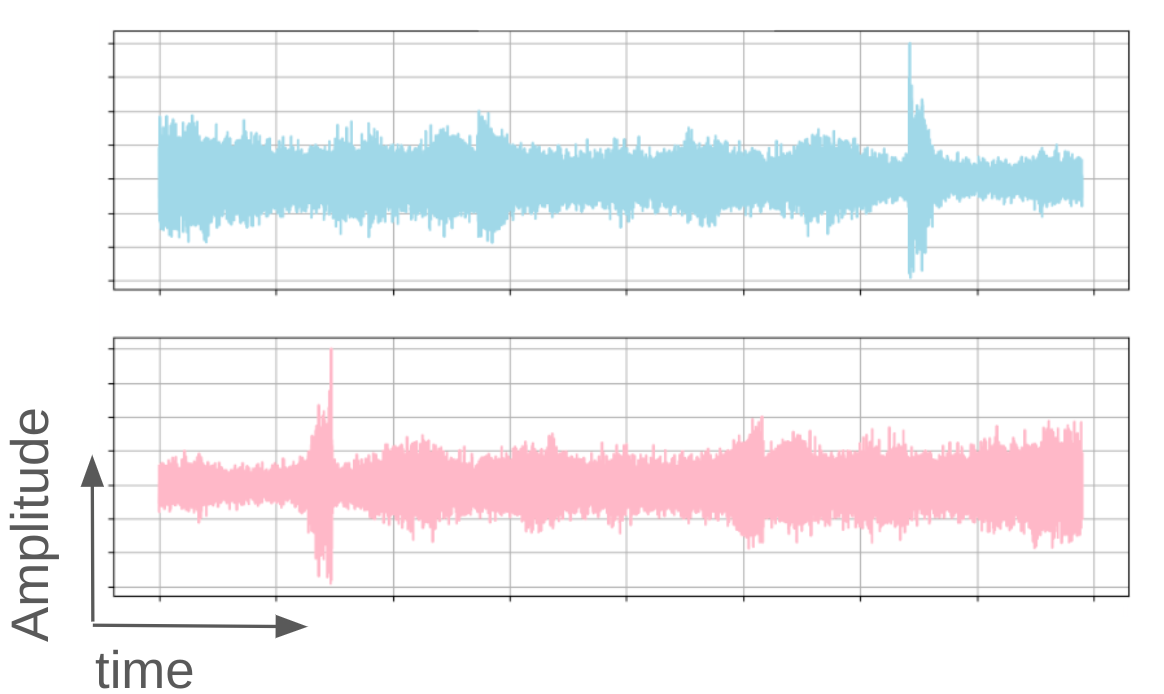}
        \caption{Reverse}
        \label{fig:reverse}
    \end{subfigure}
    \begin{subfigure}{0.24\textwidth}
        \centering
        \includegraphics[width=\textwidth]{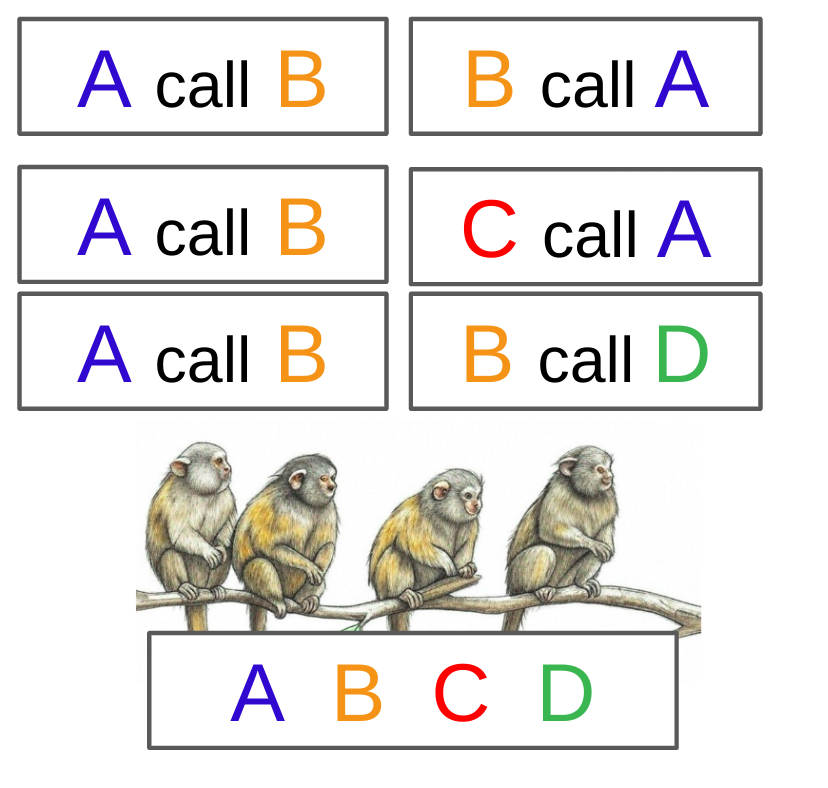}
        \caption{Phee Evals}
        \label{fig:phee}
    \end{subfigure}
    
    \caption{Our uLM evaluation tasks. The original samples are positioned above, while the pseudo-vocalizations are placed below.(d) The second row represents the CallerChange task and below the ReceiverChange task.}
    \label{fig:banchmark}
\end{figure*}

\subsection{\method}
\newpara{Vocalization-to-Unit (V2U).} We adapt the HuBERT model, pretrained on LibriSpeech~\citep{panayotov2015librispeech}, to tokenize \monk vocalizations. HuBERT learns speech representations via a masked prediction task on continuous audio features. To obtain discrete units, we apply k-means clustering on layer activations. Our adaptation follows a two-stage process: (i) training k-means on intermediate-layer features from HuBERT, then (ii) fine-tuning HuBERT using the new quantizer. Final units are extracted from the first layer of the fine-tuned model trained on \datapre.

\newpara{unit-LM (uLM).} We use the vanilla Transformer architecture within a conventional language modeling framework. Unlike previous speech-related studies that found performance improvements by removing sequential unit repetitions \citep{kharitonov2021text}, we discovered that unit length conveys important information. As a result, we preserve repeated units rather than collapsing them.

\newpara{Unit-to-Vocalization(U2V).} To reconstruct \monk vocalizations from the generated unit sequences, we utilize a unit vocoder, following the approach of \citet{polyak21_interspeech}. Since \datapre lacks speaker identity annotations, we exclude speaker embeddings and F$0$ information, relying only on vocalization tokens, obtained  by V2U.

\subsection{uLM evaluation}
\label{sec:eval_method}
Drawing inspiration from established evaluation frameworks for text and speech-based large language models (LLMs) \citep{on_generative}, we develop a set of zero-shot evaluation tasks.

We follow evaluation methods from previous research on lexical and syntactic modeling in SpeechLMs, specifically sWUGGY and sBLIMP \citep{nguyen2020lwuggy}. In sWUGGY, the model receives pairs of utterances—one with a real word and the other with a phonotactically plausible non-word—and is assessed based on its ability to assign a higher probability to the real word. Similarly, sBLIMP evaluates the model using paired speech segments, where one contains a grammatically correct sentence and the other an ungrammatical one.

To adapt these methods to our setting, we create pairs of \monk speech (positive samples) and pseudo-\monk speech (distractor samples) by systematically modifying natural \monk call sequences. Specifically, we generate distractor samples using four approaches: (i) \textit{Shuffle} (Figure \ref{fig:shuffle}), where we randomly reorder the sequence of \monk calls within the same segment; (ii) \textit{Reversal} (Figure \ref{fig:reverse}), where we reverse the entire audio segment; and (iii) \textit{Concat}~(Figure \ref{fig:concat}), where we combine the first half of one audio segment (A) with the second half of another segment (B). Specifically, we take two distinct call segments, each containing six calls, split them at their midpoints, and merge the first half of one segment with the second half of another. (iv)~\textit{Phee eval} (Figure \ref{fig:phee}), where \dataft is divided into call-and-response pairs, with the call originating from one \monk and the response from another. To generate pseudo call-response pairs, we keep the original call unchanged but modify the response. This is done by either changing the responder's identity (CallerChange) or altering the intended recipient (ReciverChange), i.e., who is calling versus whom the call is directed to. See Figure~\ref{fig:banchmark} for a visual representation of all four evaluation tasks. 

Finally, we evaluate the uLM’s capability to assign a higher probability to the real \monk segment over its pseudo-\monk counterpart. Higher accuracy suggests that the model effectively captures meaningful structures in the unit sequences, allowing it to differentiate authentic sequences from artificially modified ones. It is important to note that while the first three evaluation tasks are artificially constructed, they offer a quick and efficient method for assessing the uLM's ability to model the likelihood of \monk vocalizations. The final evaluation task (Phee eval) is particularly noteworthy and reliable because (i) it is based on manually labeled data, and (ii) previous research indicates that the likelihood of a pseudo pair naturally occurring in the training data is extremely low~\cite{oren_vocal_2024}.

%% file: sections/05_setup.tex
\section{Experimental Setup}
\label{sec:setup}
\newpara{Dataset.} 
\datapre was then divided into training ($80\%$), validation ($10\%$), and test ($10\%$) sets. Each segment was $10$ seconds long and contained as many \monk calls as possible. Overall, the dataset includes $216K$ samples. Dataset statistics are detailed in Table \ref{tab:general data}. For \dataft, a total of $56$ call-response pairs were identified, in which one \monk calls another, and the recipient responds, with gaps of up to $10$ seconds between calls. Using these pairs, we generated approximately $600$ augmented sequences to create the evaluation task, as outlined in Section~\ref{sec:eval_method}. Full details about the dataset can be found on Section~\ref{sec:data_setup}.

\newpara{Training configuration.} All training configurations including hyper-parameters, compute resources, implementation details, etc. can be found on Section~\ref{sec:train_conf} in the Appendix.

%% file: sections/06_results.tex
\section{Results}
\label{sec:res}
\subsection{Main results}
We start by evaluating \method using the Shuffle, Concat, and Reversal tasks, along with reporting uLM Perplexity (PPL). We compare the performance of the proposed method against uLMs trained on \datapre using vocalization units from three V2U quantizers: (i) S-Hu, where both HuBERT and k-means were trained on speech-only data; (ii) S-mHu, where HuBERT was trained on speech, but k-means was trained on \datapre; and (iii) the proposed method where both HuBERT and k-means were trained entirely on \datapre. 

Results are presented in Table~\ref{tab:hubert_performance}. We observe consistent improvements across all evaluation tasks as training progresses, indicating that gradually adapting the \ac{GSLM} pipeline to \monk vocalizations enhances its ability to capture meaningful patterns and more effectively model data likelihood.

\begin{table}[t!]    
    \centering
    \resizebox{\linewidth}{!}{
    \begin{tabular}{lcccc}
        \toprule
        & \textbf{Shuffle} & \textbf{Concat} & \textbf{Reversal} & \textbf{PPL} \\
        \midrule
        S-Hu  & 67.75  & 78.96 & 83.29 & 2.02 \\
        S-mHu  & 71.57  & 70.31 & 87.33 & 2 \\
        \textbf{\method}  & \textbf{84.84} & \textbf{79.94} & \textbf{90.45} & \textbf{1.78} \\
        \bottomrule
    \end{tabular}}
    \caption{Shuffle, Reversal and Concat tasks performance across
different models. All baseline models were trained on units extracted from the 9th layer.\label{tab:hubert_performance}}
\end{table}

Next, we evaluate \method using the Phee eval evaluation task, as described in Section~\ref{sec:eval_method}, on \dataft. Similar to the previous evaluations, we compare \method's performance against S-Hu and S-mHu. The results, presented in Table \ref{tab:phee_calls}, show that S-Hu, trained exclusively on human speech, performs at near-random levels ($\sim50\%$) on the ReceiverChange task. Adapting the k-means module to model \monk vocalizations (S-mHu) improves performance, achieving $\sim59\%$ on both ReceiverChange and CallerChange. \method significantly outperforms both baselines considering both ReceiverChange and CallerChange tasks.
\begin{table}[t!]    
    \centering
    \resizebox{0.9\linewidth}{!}{
    \begin{tabular}{lcc}
        \toprule &
         \textbf{ReceiverChange} & \textbf{CallerChange} \\
        \midrule        
        S-Hu  & 50.76 & 59.67 \\
        S-mHu & 58.77 & 59.99 \\
        \textbf{\method} & \textbf{65.26} & \textbf{71.51} \\
        \bottomrule
    \end{tabular}}
    \caption{Phee evaluation metrics performance across different models.\method significantly outperforms both baselines. \label{tab:phee_calls}}
\end{table}

\begin{figure}[t!]
    \centering
    \includegraphics[width=\linewidth]{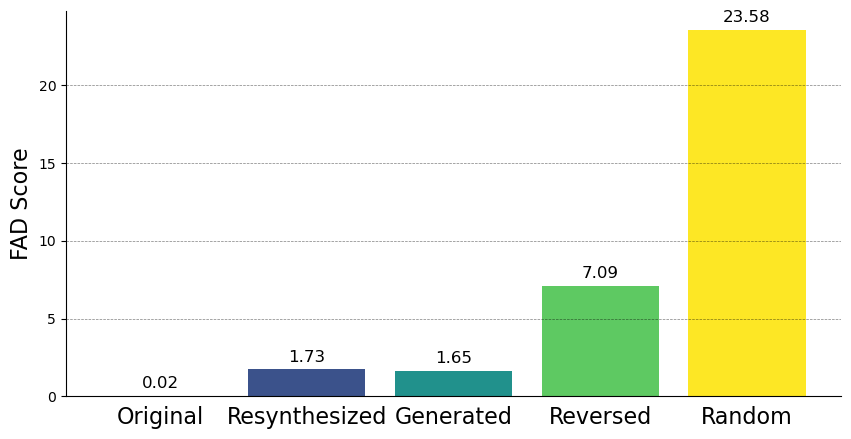}
    \caption{FAD scores for different audio manipulations (lower is better). The x-axis shows subsets compared to an original reference set. See Appendix \ref{appendix:FAD} for details.}
    \label{fig:fad}
\end{figure}
Lastly, we evaluate the quality of our generated \monk vocalizations using the Fréchet Audio Distance (FAD) \citep{kilgour2019frechetaudiodistancemetric}, computed with a VGGish encoder~\citep{hershey2017cnn}. FAD is a perceptual similarity metric that measures the distributional distance between real and generated audio embeddings, lower FAD scores indicate greater similarity to real vocalizations. As shown in Figure \ref{fig:fad}, both the generated and resynthesized vocalizations exhibited closely match, low FAD scores, significantly lower than those of reversed and random audio samples. These findings underscore the naturalness of our generated vocalizations in comparison to real data.  
We provide examples of resynthesized and generated \monk vocalizations visually in Figure~\ref{fig:spectrograms}, and audio in \href{https://pages.cs.huji.ac.il/adiyoss-lab/GmSLM/}{link}.

\subsection{Model evaluation}
\label{sec:generalization}
To further assess the quality and generalizability of the learned representations, we conducted supervised classification experiments using the final uLM outputs from our GmSLM model. Specifically, we trained simple classifiers on the open-source \monk vocalization dataset, \datalab~\cite{sarkar2023}, which contains 11 vocalization types recorded from 10 individual marmosets (licensed under Creative Commons Attribution 4.0). We evaluated two classification tasks: predicting the vocalization type and identifying the individual speaker. As shown in Table~\ref{tab:classification_scores_for_GMSLM}, the GmSLM representations achieved high performance across both tasks, with an F1 score of 90.72 for vocalization type classification and 90.12 for speaker identity classification. These results demonstrate that the model captures rich, discriminative information about both vocal content and speaker identity. Importantly, they highlight the model's ability to generalize beyond its training objective, and suggesting that the model generalizes well across different marmoset setups, supporting its applicability in broader downstream tasks that may serve as baselines.
Further implementation details regarding the classifiers are provided in Appendix~\ref{appendix:classifier-details} and Figure \ref{fig:ablation_model_comp} and Table \ref{tab:classifier_gmslsm_hubert} provide comparison of different feature representations on the same classification tasks. \method consistently outperforms all baselines.

\begin{table}[t!]
    \centering
    \resizebox{0.9\linewidth}{!}{
    \begin{tabular}{lccc}
    \toprule
    \textbf{Task} & \textbf{Recall} & \textbf{Precision} & \textbf{F1} \\
    \midrule
 Vocalization Type & 91.96 & 90.14  & 90.72  \\
    Speaker Identity & 90.25 & 90.03  & 90.12  \\
    \bottomrule
    \end{tabular}}
    \caption{Classification performance scores. Results report the average across four different random seeds, std is smaller than 0.001 in all cases.}
\label{tab:classification_scores_for_GMSLM} 
\end{table}

\subsection{Ablation study}
\label{sec:ablation}

We conduct an ablation study to examine the impact of (i) the number of clusters and unit duplication (ii) the choice of HuBERT layer used for feature extraction when training the uLM and (iii) the impact of using SSL features and a Transformer uLM versus simpler alternatives in the GmSLM model.

\newpara{The effect of number of clusters and unit duplications.} Table~\ref{tab:base_results} presents the results for the Shuffle, Concat, and Reversal evaluation tasks. We assess the proposed method using both $50$ and $100$ units, with and without unit deduplication. When comparing the number of units, results indicate that $50$ clusters perform best on the Shuffle evaluation task and yield comparable performance on the other evaluations. As a smaller vocabulary size results in shorter sequences and more efficient uLM training, we prioritize a vocabulary size of $50$. When examining unit deduplication, results show that preserving unit repetitions improves performance across all tasks, suggesting that unit length is beneficial in \monk vocalization modeling. This contrasts with spoken language, where repetitions hurt performance \citep{on_generative}. We hypothesize this stems from structural differences and fewer vocalizations in marmosets. This aligns with Section~\ref{sec:analysis}, where low unit purity suggests meaning arises from context rather than individual units.

\begin{table}[t!]
    \centering
    \resizebox{\linewidth}{!}{
    \begin{tabular}{lcccccc}
        \toprule
        \textbf{Dedup Units} & \textbf{Clusters} & \textbf{Shuffle} & \textbf{Concat} & \textbf{Reversal} \\
        \midrule
        V  & 100 & 62.64  & 77.29  & 75.55  \\
        V  & 50  & 64.87  & 77.68  & 75.59  \\
        \textbf{X}  & \textbf{50}  & \textbf{67.75}  & \textbf{78.96}  & \textbf{83.29} \\
        \bottomrule
    \end{tabular}}
    \caption{Number of clusters and duplications analysis.\label{tab:base_results}}    
\end{table}

\begin{figure}[t!]
    \centering
    \includegraphics[width=\linewidth]{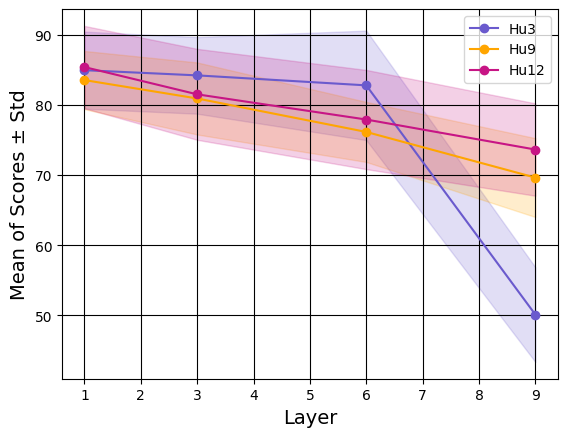}
    \caption{Mean and standard deviation results of uLM trained on tokens derived from various HuBERT models $Hu<n>$, where \( n \in \{3, 9, 12\} \), indicating different layer supervision). The x-axis represents layers used for discretized token creation.
    \label{fig:mean result for teachers and layers}
    }    
\end{figure}

\newpara{The effect of layer selection.} Next, we analyze the impact of selecting different layers for quantization and uLM training. We extract and quantize features from layers $1, 3, 6$, and $9$, training a separate uLM for each. To ensure the results are consistent across different encoders, we repeat this process using three HuBERT models, each differing in the layer used for teacher supervision. For readability, we report the average results and standard deviations for the Shuffle, Concat, and Reversal evaluation tasks, as shown in Figure~\ref{fig:mean result for teachers and layers}. The findings indicate that uLM performance declines as deeper layers are used. We hypothesize that this occurs because \monk vocalizations are more acoustic than semantic (contextualized), making it easier for the uLM to capture fine-grained acoustic variations in earlier layers rather than in deeper layers.

\begin{figure}[t!]
    \centering
    \includegraphics[width=\linewidth]{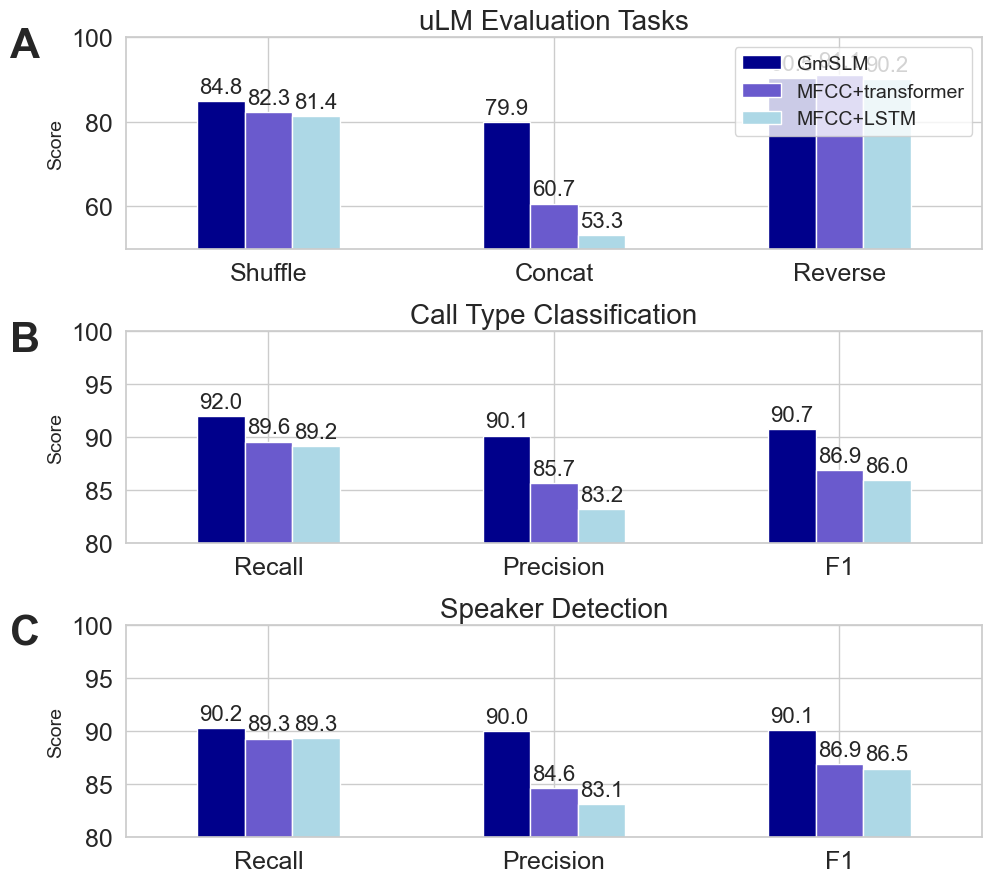}
    \caption{GmSLM vs. Simpler Models
    }
    \label{fig:ablation_model_comp}
\end{figure}

\newpara{The effect of model complexity.}
To evaluate whether \method is overly complex for modeling \monk vocalizations, we compare it against simpler alternatives varying both input features and uLM architecture: (i) MFCC v2U + Transformer uLM and (ii) MFCC v2U + LSTM uLM. Results (Figure~\ref{fig:ablation_model_comp}A) show that \method outperforms both, particularly on the Concat task requiring cross-call context. Additionally, we experimented with a custom filterbank tailored to the energy profile of \monk vocalizations. This alternative achieved performance comparable to the MFCC-based setup (Appendix~\ref{appendix:monk_filterbank}).
 Classifier evaluations (trained as described in Section~\ref{sec:generalization})  further confirm that \method representations significantly surpass those from the simpler models, underscoring the effectiveness of SSL features and Transformer modeling for this data (Figure~\ref{fig:ablation_model_comp}B,C). 
 These results suggest that SSL-based representations provide richer features, particularly for capturing contextual aspects of vocalizations and for classification performance, and that the advantage of HuBERT shallower layers does not indicate a lack of semantic properties (contextualized information) but rather an effective balance between acoustic and semantic information.(Full training details for the LSTM model are provided in Appendix~\ref{appendix:LSTM-details}).

\subsection{Analysis}
\label{sec:analysis}

A central question in \monk vocalization research is the role of context in communication~\citep{eliades2017marmoset}. To explore this, we systematically limited the model’s attention to a fixed context length by masking all tokens beyond a threshold (set to $-\infty$), effectively removing them from the softmax.We tested context spans of $500$, $400$, $300$, $200$, and $50$ tokens, with and without preserving the first token or first five tokens—known to influence attention~\citep{sun2024massive,oren2024transformers}. Table~\ref{tab:context_analysis} reports results for the Shuffle, Concat, and Reversal tasks.

Without preserving early tokens, performance matched the full-context setup only at $500$ tokens ($\sim10$ seconds). When keeping the first token, similar performance persisted down to $300$ tokens, suggesting that essential contextual information in \monk vocalizations spans $\sim6$ seconds.

\begin{table}[t!]   
    \centering
    \resizebox{\linewidth}{!}{
    \begin{tabular}{l l ccc}
        \toprule
        \makecell{\textbf{Context} \\ \textbf{Length}} & \makecell{\textbf{First} \\ \textbf{Tokens}}
        & \makecell{\textbf{Shuffle}} & \textbf{Concat} & \textbf{Reversal} \\
        \midrule
        - & - & 83.15 & 79.6  & 93.44 \\
        \midrule
        \multirow{3}{*}{500} & 0 & 83.15 & 79.54 & 93.44 \\
                             & 1 & 83.15     & 79.54     & 93.44     \\
                             & 5 & 83.15    & 79.54    & 93.42     \\
        \midrule
        \multirow{3}{*}{400} & 0 & 81.54 & 79.42 & 85.72 \\
                             & 1 & 82.79    & 79.42     & 93.42     \\
                             & \textbf{5} & \textbf{83.04}    & \textbf{79.54}     & \textbf{93.42}      \\
        \midrule
        \multirow{3}{*}{300} & 0 & 75.14 & 79.01 & 73.5 \\
                             & 1 & 81.54     & 78.68     & 93.18     \\
                             & 5 & 82.33     & 78.85     & 93.28  \\
        \midrule
        \multirow{3}{*}{200} & 0 & 62.71 & 75.76 & 67.71 \\
                             & 1 & 79.19 & 78.23 & 92.69      \\
                             & 5 & 80.4     & 78.19    & 93.1   \\
        \midrule
        \multirow{3}{*}{50}  & 0 & 30.74 & 32.49 & 35.81 \\
                             & 1 & 68.95 & 70.14 & 84.94     \\
                             & 5 & 71.9    & 70.69     & 88.23    \\
        \bottomrule

    \end{tabular}}
    \caption{The effect of context length on \method language modeling performance, considering the first token, the first five tokens, or no tokens at all. \label{tab:context_analysis}}
\end{table}

To better understand the role of context alongside our findings on unit repetitions, we analyzed how well uLM units align with vocalization labels. Specifically, we examined the alignment between units—obtained via $k$-means clustering over HuBERT representations—and annotated call types. On a balanced subset of \datalab, we observed a vocalization purity of 0.18 and a unit purity of 0.26, both relatively low compared to speech language models. These results support the view that unit meaning depends on context and unit length rather than being inherently discrete or standalone (Section \ref{sec:generalization}). This suggests the presence of an underlying n-gram-like distribution over units, which remains an open direction for future work. Full distributions are shown in Figures~\ref{fig:labels_distribution_of_units}, \ref{fig:Units Distribution for different calltype}, and~\ref{fig:wide_plot}. Analysis of pre-softmax attention matrices (Figure~\ref{fig:attention}) revealed no consistent patterns aligned with call labels.

%% file: sections/02_related.tex
\section{Related work}
\label{sec:related}

\newpara{Textless NLP} was introduced by \citet{on_generative}, showing how raw speech can be used to build \ac{GSLM} systems. \citet{kharitonov2021text} extended this with multi-stream speech language models (SLMs) that combine pseudo-text with prosodic features, opening new directions in spoken language modeling. Later works improved performance by initializing SLMs from text LMs \cite{twist}, framing tasks like emotion \cite{kreuk2021textless} and speaking style conversion \cite{maimon2022speaking} as discrete translations, and jointly modeling dialogue \cite{nguyen2022generative}. \citet{borsos2022audiolm} proposed a cascade of LMs for semantic and acoustic tokens, enabling high-quality, speaker-consistent speech generation \cite{wang2023neural, kharitonov2023speak}. Semantic units from these models correlate with phonemes \cite{sicherman2023analysing}, while others adapted this approach for speech translation \cite{popuri2022enhanced, peng2024mslm}. Additional advances include augmenting text LMs with speech for QA \cite{nachmani2023spoken}, state-space SLMs for long-context modeling \cite{park2024long}, and preference-tuned SLMs using LLM feedback \cite{lin2024alignslm, rafailov2024dpo}.

\newpara{Animal language.} Traditional analysis of marmoset vocalizations (e.g., caller identity, call type) relied on ML models trained on acoustic features \cite{turesson2016, verma2017discovering}. More broadly, ML, and particularly deep learning—has been used across various species for vocalization analysis, considering birds \cite{kahl2021birdnet,ghani2023global}, dogs \cite{huang2023transcribing}, mice~\cite{coffey2019deepsqueak}, call classification \cite{zhang2018}, and nonhuman primates \cite{pellegrini2021,oikarinen2019}, including marmosets \cite{uesaka2023marmoset}.

Recent advances in self-supervised learning (SSL), originally developed for human speech, have enabled large-scale bioacoustic learning from unlabeled data, supporting tasks like birdsong detection \cite{saeed2021}, event detection \cite{bermant2022}, and multi-species classification \cite{hagiwara2023aves}. SSL has been used for phonetic and lexical discovery in dogs \cite{wang2024doghubert, abzaliev2024dogbarkdecodingleveraging}, caller/call-type classification in marmosets \cite{sarkar2023}, and gibbon identity recognition \cite{cauzinille2024gibbon}. These representations have also been leveraged to infer animal emotions and health from vocalizations \cite{manikandan2024decoding}.
Parallel to our work, \citet{kobayashi2025finchgpttransformerbasedlanguage} proposed FinchGPT—a Transformer-based model trained on birdsong transcripts—demonstrating its capacity to capture long-range dependencies in syllable sequences, consistent with our findings.

%% file: sections/07_conclusion.tex
\vspace{-0.1cm}
\section{Conclusion}
\label{sec:con}
\vspace{-0.1cm}
We introduce \textbf{Generative Marmoset Spoken Language Modeling} (\method), a novel pipeline for modeling, generating, and evaluating \monk vocalizations without labeled supervision. Our approach bridges nonhuman primates vocal communication and modern generative language models. Through zero-shot evaluations, \method effectively captures structural properties of \monk vocalizations, distinguishing authentic from artificial utterances. It outperforms human-speech-only baselines and achieves high performance on downstream tasks.

Layer selection and model complexity analyses show that SSL-based representations provide richer features, with the first layer performing best—likely reflecting an optimal balance between acoustic and semantic information in Marmoset vocalizations. Contextual analysis indicates key communicative information spans $\approx$six seconds.

Our work introduces a new direction for unsupervised spoken language modeling beyond human speech. We position \method as a foundational tool for studying vocal communication in species with limited annotations, fostering interdisciplinary research at the intersection of computational linguistics, bioacoustics, and neuroscience.

%% file: Figures/spect_gen_rsynt_examples.tex
\begin{figure*}[t]
  \centering
  \includegraphics[width=\textwidth]{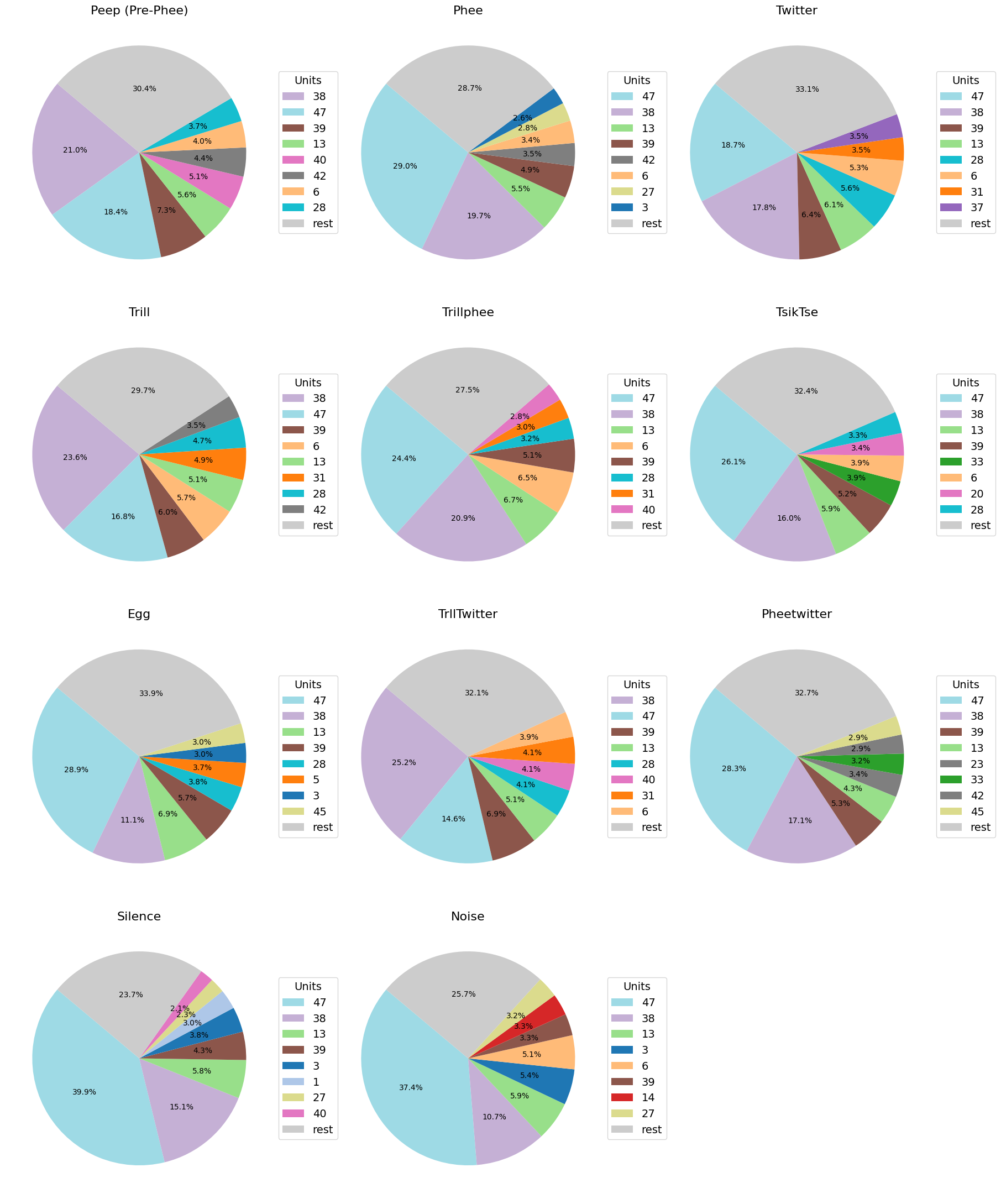}
  \caption{Units distribution for different call-types given from \datalab. Analysis is based on a balanced subset from the dataset, where each call-type occur equaly often and was chosen randomly. The call type names (Peep, Phee etc.) represents known vocal patterns in the marmoset vocalizations.}
  \label{fig:labels_distribution_of_units}
\end{figure*}

\begin{figure*}[t]
  \centering
  \includegraphics[width=\textwidth]{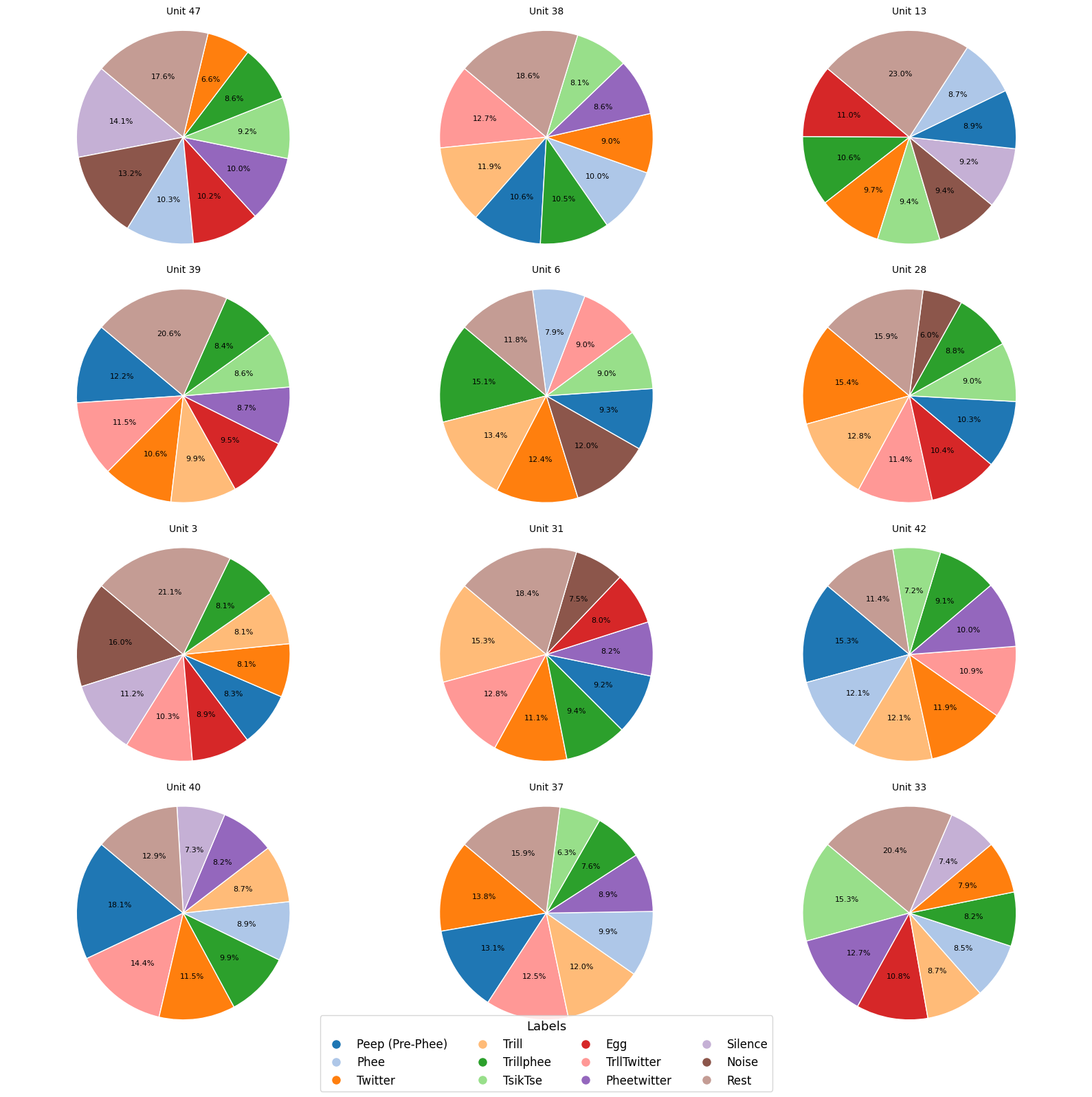}
  \caption{Vocalization Type Labels distribution for the most frequent HuBERT units. The analysis is based on a balanced subset of \datalab dataset, where each call-type occur equaly often and was chosen randomly. The call type names (Peep, Phee etc.) represents known vocal patterns in the marmoset vocalizations. HuBERT units are given from the 9th layer of a the marmoset-HuBERT model used for GmSLM}
  \label{fig:Units Distribution for different calltype}
\end{figure*}

\begin{figure*}[t]
  \centering
  \includegraphics[width=\textwidth]{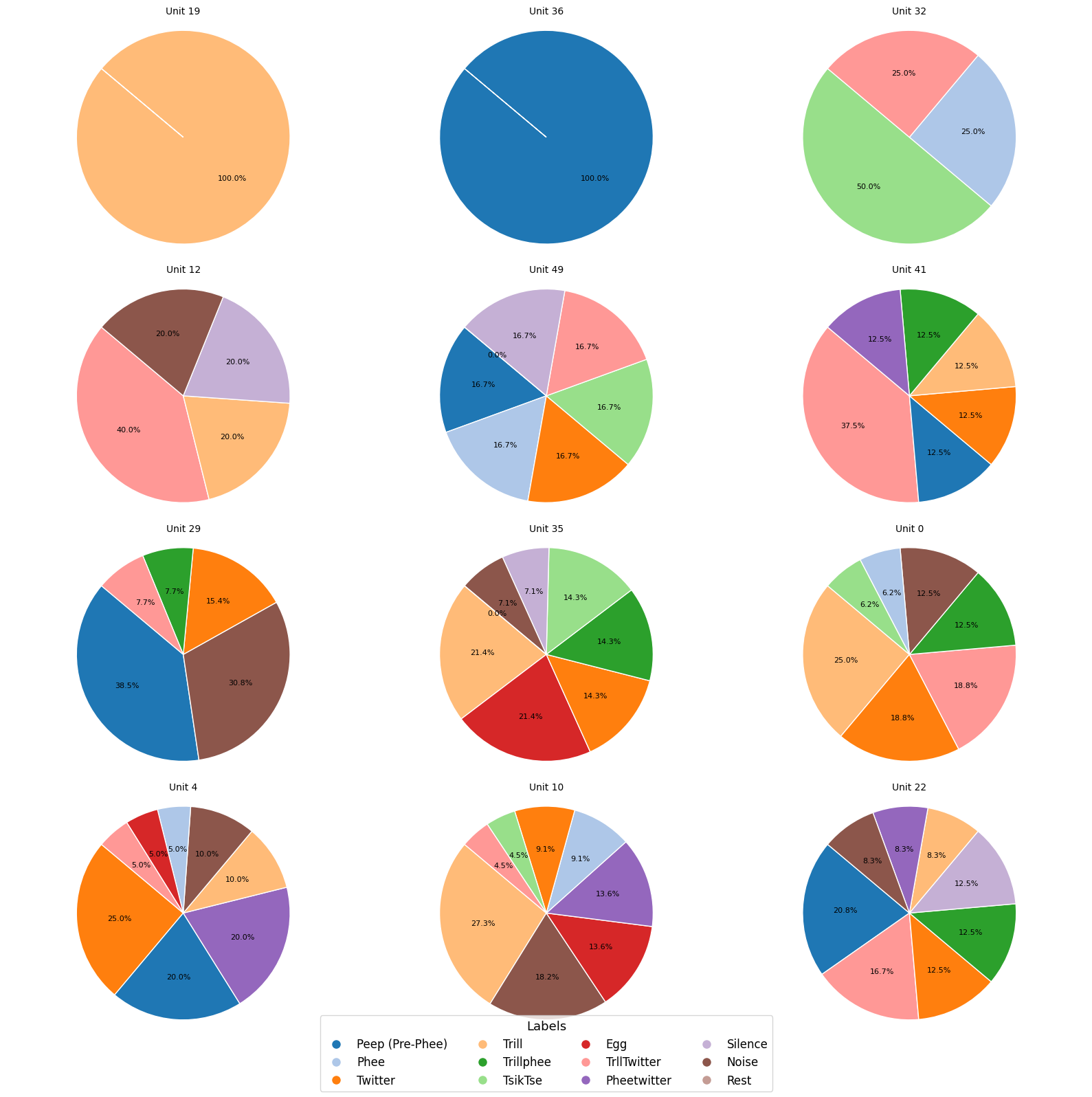}
    \caption{Vocalization Type Labels distribution for the less frequent HuBERT units. The analysis is based on a balanced subset of \datalab dataset, where each call-type occur equaly often and was chosen randomly. The call type names (Peep, Phee etc.) represents known vocal patterns in the marmoset vocalizations. HuBERT units are given from the 9th layer of a the marmoset-HuBERT model used for GmSLM}
  \label{fig:wide_plot}
\end{figure*}

\begin{figure*}[t!]
    \vspace{-1cm} 
    \centering
    \begin{subfigure}{\textwidth}
        \centering
        \includegraphics[width=\textwidth]{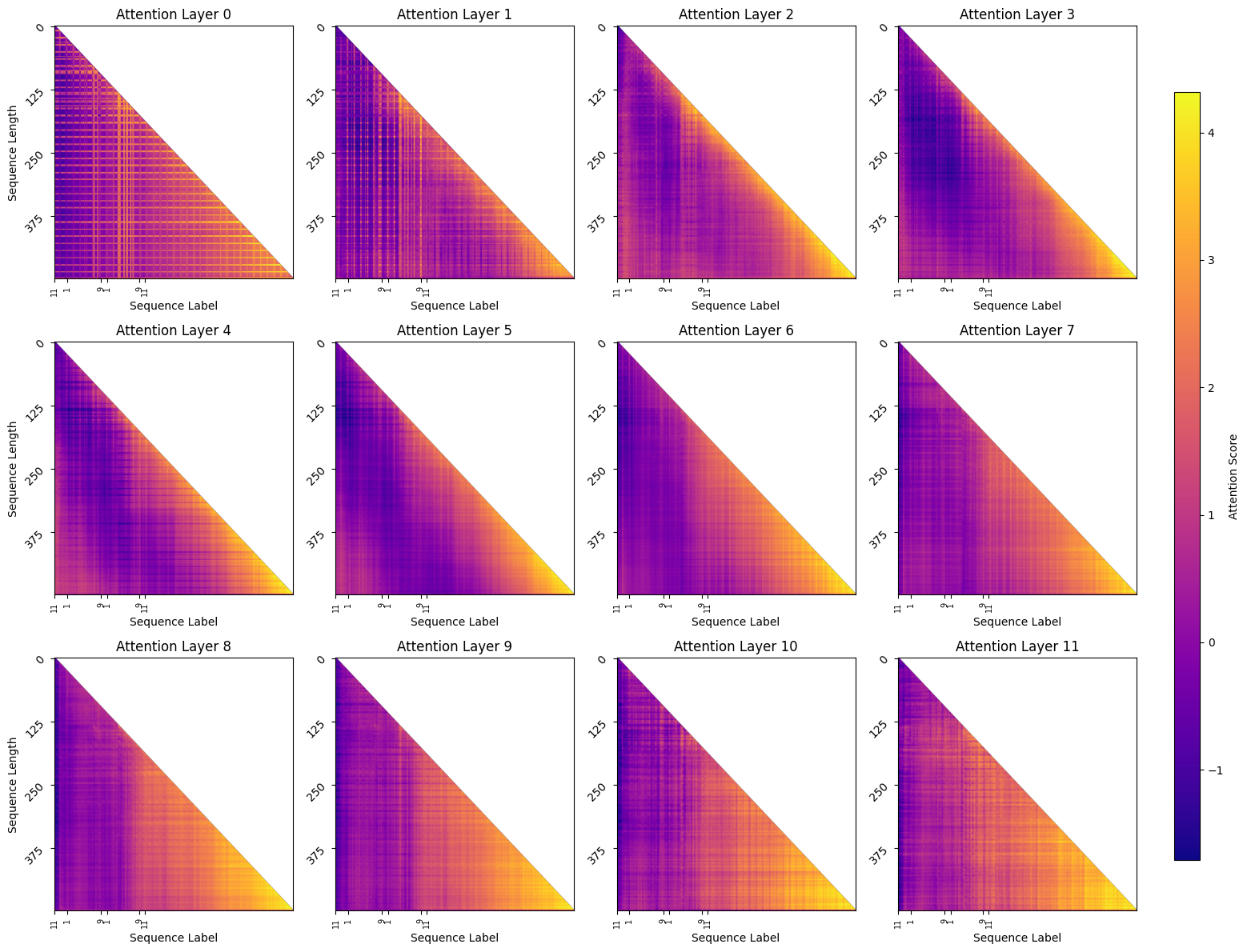}
        \label{fig:spect_example}
    \end{subfigure}


    \begin{subfigure}{\textwidth}
        \centering
        \includegraphics[width=\textwidth]{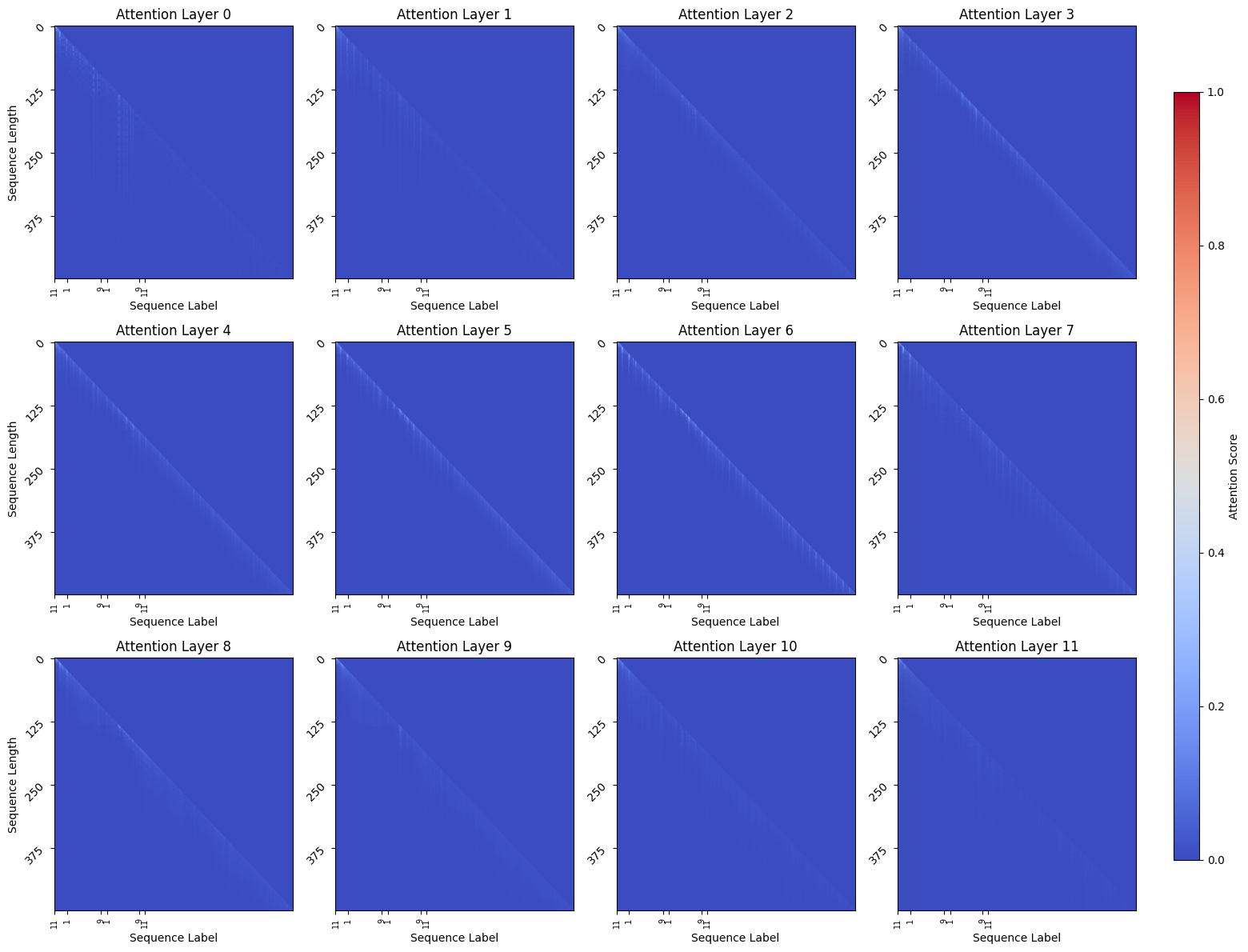}
        \label{fig:spect_example_2}
    \end{subfigure}
    \vspace{-1.3cm} 

    \caption{Attention map averaged across different heads of a sample from \datalab before and after applying softmax. the x-axis represents the start time of call labels where '11' represents silence, '1' is a Phee call, '9' is a 'Pheetwitter' call. We did not observe a clear relationship between attention patterns and the asigned labels.}
    \label{fig:attention}
\end{figure*}

\begin{figure*}[t!]
    \vspace{-1cm} 
    \centering
    \begin{subfigure}{\textwidth}
        \centering
        \includegraphics[width=\textwidth]{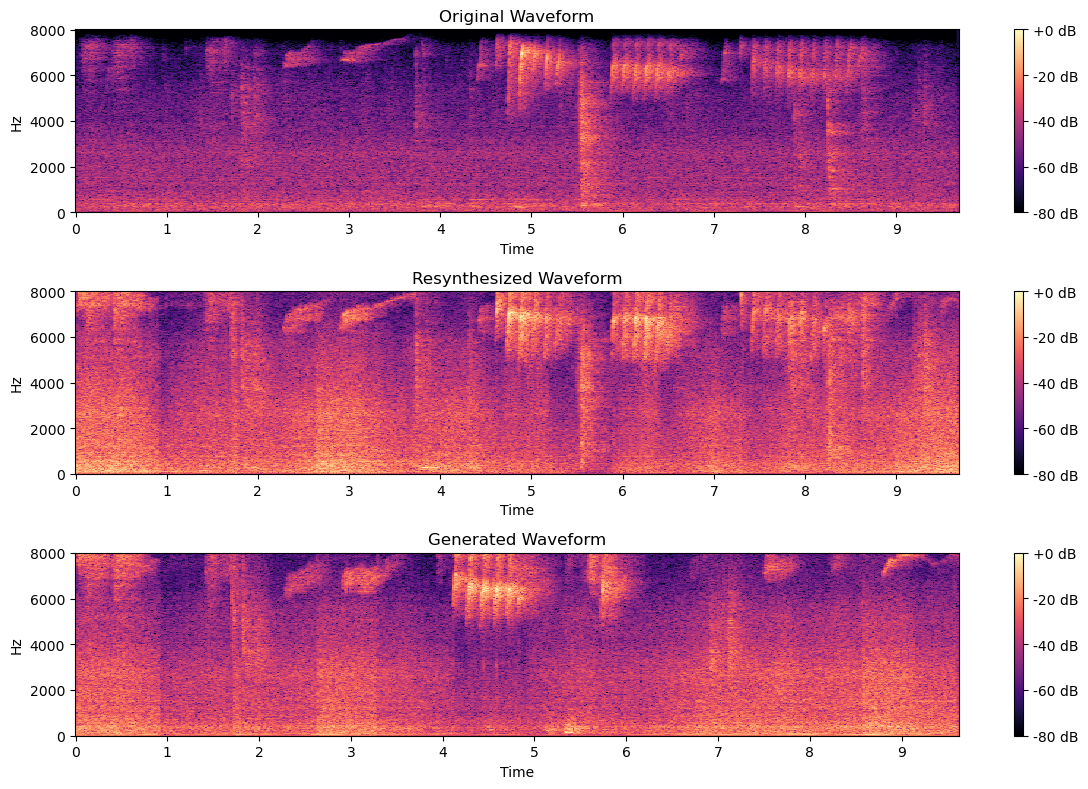}
        \label{fig:spect_example}
    \end{subfigure}

    \vspace{0.3cm} 

    \begin{subfigure}{\textwidth}
        \centering
        \includegraphics[width=\textwidth]{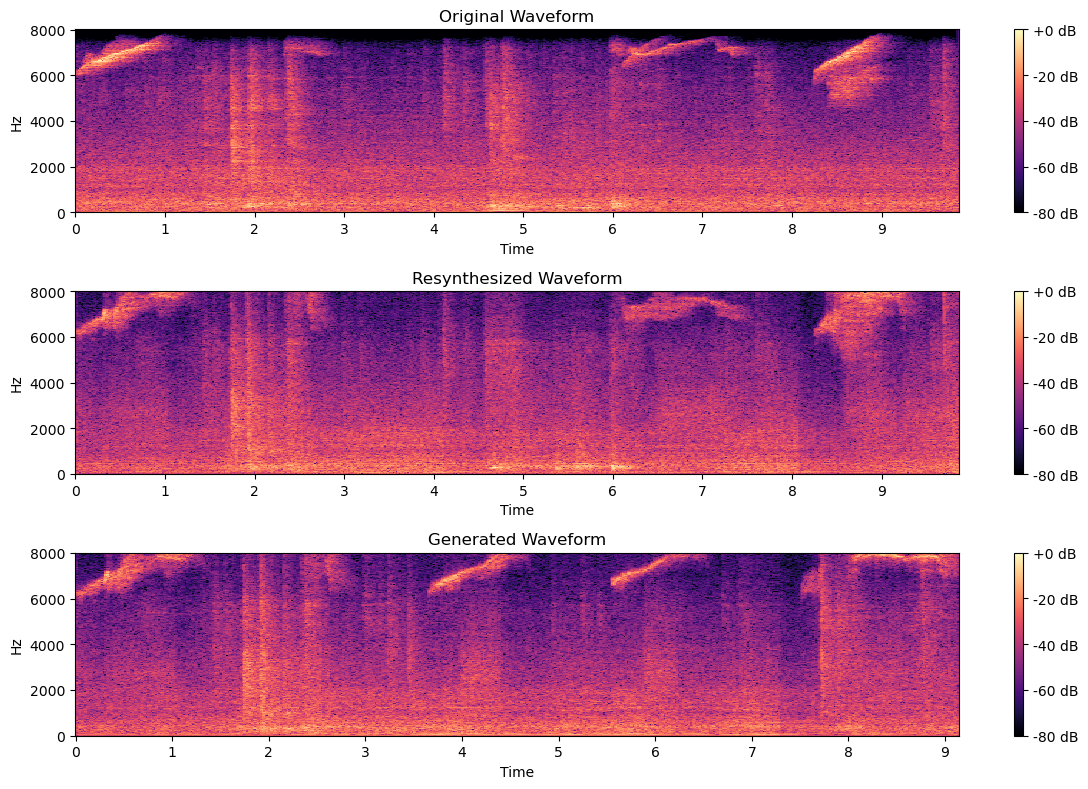}
        \label{fig:spect_example_2}
    \end{subfigure}

    \caption{Spectrograms of an original \monk vocalization, its resynthesized version, and a generated version produced using \method, conditioned on the first 3 seconds as a prompt.}
    \label{fig:spectrograms}
\end{figure*}